\documentclass{article}
\usepackage{graphicx}
\usepackage{soul}

\usepackage{cleveref}

\usepackage{siunitx}

\begin{document}
\title{Training dataset and dictionary sizes matter in BERT models: the case of Baltic languages}
%\titlerunning{Dataset and dictionary sizes matter in BERT models}
\author{Matej Ulčar,
Marko Robnik-Šikonja\\
University of Ljubljana, Faculty of Computer and Information Science,\\ Ljubljana, Slovenia\\
\{matej.ulcar,marko.robnik\}@fri.uni-lj.si
}
%\author{First Author \and Second Author}
%
%\authorrunning{M. Ulčar and M. Robnik-Šikonja}
%\institute{
%University of Ljubljana, Faculty of Computer and Information Science,\\ Ljubljana, Slovenia\\
%\email{\{matej.ulcar,marko.robnik\}@fri.uni-lj.si}}
%\institute{Anonymous Institute}
%
\maketitle              % typeset the header of the contribution
\begin{abstract}
Large pretrained masked language models have become state-of-the-art solutions for many NLP problems. While studies have shown that monolingual models produce better results than multilingual models, the training datasets must be sufficiently large. We trained a trilingual LitLat BERT-like model for Lithuanian, Latvian, and English, and a monolingual Est-RoBERTa model for Estonian. We evaluate their performance on four downstream tasks:  named  entity  recognition, dependency parsing, part-of-speech tagging, and word analogy. To analyze the importance of focusing on a single language and the importance of a large training set, we compare created models with existing monolingual and multilingual BERT models for Estonian, Latvian, and Lithuanian. The results show that the newly created LitLat BERT and Est-RoBERTa models improve the results of existing models on all tested tasks in most situations. 

%Natural language processing  \and BERT \and transformers \and Estonian \and Latvian \and Lithuanian.}
\end{abstract}

\section{Introduction}
Large pretrained language models, based on transformers \cite{Vaswani2017} present the current state of the art in solving natural language processing tasks. These models are trained on large corpora in a self-supervised fashion and are able to learn basic language principles. Because of this and their ability to quickly and successfully adapt to a wide variety of tasks, they have also been (somewhat impetuously) called foundation models \cite{bommasani2021opportunities,marcus2021aifoundation}.

The largest, most complex transformer based models have been trained only for the languages with most resources, mainly English, or in a massive multilingual fashion, covering 100 or more languages in a single model. Two examples of such models are GPT-3 \cite{Brown2020GPT3short}, and T5 \cite{raffel2020exploring}. For most languages, however, there are neither enough training data nor compute resources to train such complex models.  Most monolingual models share the architecture of the BERT-base model \cite{Devlin2019}, e.g. RuBERT \cite{kuratov2019adaptation} for Russian, FinBERT \cite{virtanen2019multilingual} for Finnish, KB-BERT \cite{swedish-bert} for Swedish, or RobeCzech \cite{straka2021robeczech} for Czech.

It has been shown \cite{virtanen2019multilingual} that monolingual models outperform massive multilingual models, like multilingual BERT (mBERT) \cite{Devlin2019} or XLM-RoBERTa (XLM-R) \cite{conneau2019unsupervised}. However, as we show, a good monolingual model needs to be trained on a sufficiently large corpus, and focusing on a single language alone is not sufficient for good performance. 

In this work, we introduce two new large models and make them publicly available. Est-RoBERTa\footnote{https://doi.org/10.15155/9-00-0000-0000-0000-00226L} is a monolingual Estonian model, while LitLat BERT\footnote{http://hdl.handle.net/20.500.11821/42} is a trilingual model, trained on Lithuanian, Latvian, and English corpora. This enables us to compare these models with existing smaller monolingual models. We show that the size of training data matters and that when data is scarce, additional corpora from a  similar language can aid in the model's performance. %multilingual model trained on few languages can significantly outperform a monolingual model.

The paper is divided into five sections. In \Cref{relatedWork}, we present related Lithuanian, Latvian, and Estonian BERT models. In \Cref{sec:newmodels}, we present dataset and training for creation of new  models. We evaluate the new models and compare them with existing ones in \Cref{sec:evaluation}. We draw conclusions and present ideas for further work in \Cref{conclusions}.

\section{Existing models for Estonian, Latvian, and Lithuanian} 
\label{relatedWork}

Latvian and Lithuanian are Baltic languages, belonging in the Indo-European language family. Estonian belongs in the Uralic language family and is not related to the other two. However, all three languages share a few common properties. They are all morphologically complex and they are spoken by a relatively small number of speakers, thus few resources are available. 

The massive multilingual models, mBERT~\cite{Devlin2019} and XLM-R~\cite{conneau2019unsupervised} support all three languages. Estonian is also covered by the multilingual FinEst BERT~\cite{ulcar2020xlbert}, which was trained on Estonian, Finnish, and English corpora. Few monolingual models have been trained for Estonian, Latvian, and Lithuanian.  EstBERT~\cite{tanvir2020estbert} is a monolingual Estonian model, trained on Estonian National Corpus 2017, which consists mostly of web-crawl texts and includes also newspaper texts and Estonian Wikipedia. LVBERT~\cite{znotins2020lvbert} is a monolingual Latvian model, trained on various Latvian corpora, composed mostly of articles from news portals and Latvian Wikipedia. For Lithuanian we only found a LitBERTa-uncased model\footnote{\url{https://huggingface.co/jkeruotis/LitBERTa-uncased}} for which no details have been published.

In Section~\ref{sec:newmodels}, we present two new models. Est-RoBERTa is a new monolingual model for Estonian, which was trained on a larger corpus than EstBERT. LitLat BERT is a trilingual model, trained on Lithuanian, Latvian, and English. The sizes of training corpora for the existing and new models, supporting Estonian, Latvian, and Lithuanian, are displayed in Table~\ref{tab:datasetsize}.

%For Lithuanian: jkeruotis/LitBERTa-uncased which we did not evaluate yet, it has also not been published or properly described anywhere.

\section{LitLat BERT \& Est-RoBERTa}
\label{sec:newmodels}
In this section, we present the training of the two new models: monolingual Estonian Est-RoBERTa, and trilingual LitLat BERT, trained on Lithuanian, Latvian, and English.
%Est-RoBERTa and LitLat BERT. Trilingual pre-trained models FinEst BERT and CroSloEngual BERT~\cite{ulcar2020xlbert}, include two closely related languages and English. They have proved to be successful in both monolingual and cross-lingual applications. They are especially useful for knowledge transfer from a high-resource language (i.e. English), when the training data in one of the two target languages is scarce. With the same motivation, we trained LitLat BERT, combining English, Lithuanian, and Latvian.

%The existing Estonian monolingual model Est-BERT was trained on a small dataset and is performing unsatisfactory. We trained a new monolingual Estonian model Est-RoBERTa, on a much larger dataset. We opted not to train a combined Baltic-English model, including Estonian, Latvian, Lithuanian, and English, as Estonian is not related to the other two Baltic languages, and its inclusion would lower the model's ability to encode Lithuanian and Latvian.

We present the datasets used to train the models in Section~\ref{sec:datasets} and the models' architecture and training parameters in Section~\ref{sec:architecture}.

\subsection{Datasets and preprocessing}
\label{sec:datasets}
The sizes of corpora used for training LitLat BERT and Est-RoBERTa are shown in Table~\ref{tab:datasetsize}, along with dataset sizes for related BERT-like models.

\begin{table}
\caption{Training corpora characteristics of existing and new (at bottom) BERT-like models for Estonian, Lithuanian and Latvian. The training corpora sizes are in billions of tokens per language, and the dictionary sizes are in thousands of tokens. Some numbers are unknown (unk) or do not apply (marked with -). 
The numbers for mBERT are estimated from current sizes of Wikipedia (marked with *), it's not known which version of Wikipedia was used for training the model. The training size for LitBERTa is extrapolated from the given dataset size in gigabytes (marked with \dag).}%\hl{Could we get the dictionary sizes for each of the languages in multilingual models? Not possible, as some/many tokens are common for multiple languages. For language specific tokens, it's difficult to determine language on a subword level.}}
\label{tab:datasetsize}
\centering
\begin{tabular}{lcccccr}
\hline
% sloberta ~22 GB (4B tokens), litberta ~4.7 GB, xlm-r ~2.5 TB (300B tokens), finest ~20 GB (3.7B tokens), est-roberta ~16.6 GB (2.5B tokens)
Model & Estonian & Latvian & Lithuanian & Total & Dictionary\\
\hline
mBERT & 0.05* & 0.03* & 0.04* & unk & 120\\
XLM-R & 0.84 & 1.20 & 1.83 & 295.09 & 250 \\
Est-BERT & 1.15 & - & - & 1.15 & 50\\
FinEst BERT & 0.48 & - & - & 3.70 & 75 \\
LVBERT & - & 0.50 & - & 0.50 & 32\\
LitBERTa & - & - & 0.8\dag & 0.8\dag & 128\\  \hline
LitLat BERT & - & 0.53 & 1.21 & 4.07 & 84\\
Est-RoBERTa & 2.51 & - & - & 2.51 & 40\\
\hline
\end{tabular}
\end{table}

Est-RoBERTa was trained on a large Estonian corpus, consisting mostly of news articles from Ekspress Meedia, as well as Estonian part of CoNLL 2017 corpus~\cite{CoNLL2017}.

LitLat BERT was trained on large corpora from three languages: Lithuanian, Latvian, and English. Lithuanian corpora are composed of Lithuanian Wikipedia from 2018\footnote{\url{http://hdl.handle.net/11234/1-2735}}~\cite{11234/1-2735}, Lithuanian part of DGT corpus\footnote{\url{http://hdl.handle.net/11356/1197}}~\cite{steinberger2013dgt}, and LtTenTen14 corpus~\cite{jakubivcek2013tenten}. Latvian corpora consist of Latvian parts of CoNLL 2017 corpus~\cite{CoNLL2017} and DGT corpus~\cite{steinberger2013dgt}, Saeima corpus~\cite{dargis18saeima}, and news articles from Ekspress Meedia. For English corpus, we used English Wikipedia from 2018~\cite{11234/1-2735}.

The numbers in  Table~\ref{tab:datasetsize} show, that new Est-RoBERTa contains more than twice as much Estonian data compared to other models covering Estonian. The new trilingual LitLat BERT contains approximately the same amount of Latvian data as LVBERT but only half of massively multilingual XLM-R. Concerning Lithuanian data, LitLat BERT contains a third less data compared to XLM-R.

% \begin{table}[!htb]
% \caption{The training corpora for BERT models and their total size (in billions of tokens) per language. }
% \label{tab:corpora}
% \begin{tabular}{llr}
% Language	& Corpora & Size \\ \hline 
% English 	& 1 Billion Word Benchmark & 0.80 \\ 
% Estonian	& CoNLL 2017, Ekspress Meedia articles & 2.51 \\
% Latvian  	& CoNLL 2017, DGT-UD, Saeima, Ekspress Meedia articles & 0.53 \\
% Lithuanian	& Wikipedia 2018, DGT-UD, LtTenTen14 & 1.30 \\
% 	\hline 
% \end{tabular} 
% \end{table}

Before training, the corpora need to be preprocessed. First, we deduplicated all the corpora and formatted them to contain one sentence per line. We trained two sentencepiece models\footnote{\url{https://github.com/google/sentencepiece}}, one for each BERT-like model, on a subset of the whole corpora. For Est-RoBERTa this subset was randomly sampled, while for LitLat BERT, the subset had equal parts of Lithuanian, Latvian, and English text. The sentencepiece models are used to tokenize the corpora into subword tokens and encode them into numeric representations (ie. assign each token its unique id). The sentencepiece models contain \num{40000} and \num{84000} subword tokens for Est-RoBERTa and LitLat BERT, respectively. This sizes coincide with the size of the input layer of the model and represents the dictionary (also referred to as vocabulary) size of the model. The dictionary sizes for related models are shown in the rightmost column of Table~\ref{tab:datasetsize}.
%\noindent Displayed equations are centered and set on a separate
%line.
%\begin{equation}
%x + y = z
%\end{equation}
%Please try to avoid rasterized images for line-art diagrams and
%schemas. 

\subsection{Architecture and training}
\label{sec:architecture}
LitLat BERT and Est-RoBERTa both use the same architecture as RoBERTa-base model~\cite{liu2019roberta}. They are composed of 12 transformer layers, each layer has the size of 768. We used the input sequence length of 512. The training task for both models was masked token prediction. 15\% of input tokens were randomly masked, using whole-word masking, meaning that no word was partially seen and partially masked. Both models were trained using the fairseq toolkit \cite{ott2019fairseq} for 40 epochs on the corpora presented in Section~\ref{sec:datasets}. We used the Adam optimizer with parameters $\beta_1=0.9$ and $\beta_2=0.98$ and dropout value of $0.1$. We trained both models on 4 Nvidia Tesla V100 GPUs with the batch size of \num{20480} tokens per batch per GPU and gradient accumulation with steps of 32, so a total batch size of \num{2621440} tokens. The training took about 13 hours per epoch for Est-RoBERTa and about 35 hours per epoch for LitLat BERT (altogether approximately 3 weeks and 2 months, respectively). 

\section{Evaluation}
\label{sec:evaluation}
We evaluated Estonian, Latvian, and Lithuanian mono- and multilingual BERT-like models presented in Table~\ref{tab:datasetsize} on four tasks: named entity recognition (NER), dependency parsing (DP), part-of-speech (POS) tagging, and word analogy (WA). In Section~\ref{sec:evalsettings}, we describe the evaluation tasks and procedure, and in Section~\ref{sec:results}, we present the results of the evaluation.

\subsection{Evaluation settings}
\label{sec:evalsettings}
We evaluated the models on the NER and POS-tagging tasks by adding a standard classification head on top of the models and fine-tuned the models on each task. We used the classification code from Huggingface's transformers project\footnote{\url{https://github.com/huggingface/transformers}}. We fine-tuned each model for 3 epochs with the batch size 8. For the NER task, we evaluated models on the Estonian NER corpus \cite{estonian-ner}, TildeNER \cite{pinnis2012tildener} for Lithuanian, and train data of the LV Tagger \cite{paikens2012towardsner} for Latvian. We limited the scope of the task to the classification of three common named entity classes: persons, locations, and organizations.

For the POS-tagging and DP tasks we used the datasets from the Universal Dependencies project \cite{universal-dependencies}: EDT \cite{edt} for Estonian, LVTB for Latvian, and ALKSNIS for Lithuanian.
To solve the DP task, we translated the problem to a sequence labeling task \cite{strzyz-etal-2019-viable}, using arc-standard algorithm \cite{10.1162/coli.07-056-R1-07-027} for encoding the dependency trees. We fine-tuned the models for 10 epochs with a batch size 8, using a modified dep2label-bert tool\footnote{https://github.com/EMBEDDIA/dep2label-transformers}.

Traditional WA task measures the distance between word vectors in a given analogy word1 : word2 $\approx$ word3 : word4 (e.g., man : king  $\approx$ woman : queen). For contextual embeddings such as BERT, the task is modified by using a boilerplate sentence "If the word [word1] corresponds to the word [word2], then the word [word3] corresponds to the word [word4]."  We masked [word2] and attempted to predict it using masked token prediction. 
As the source of analogies, we used the multilingual culture-independent word analogy dataset \cite{ulcar-2020-multilingual}. We consider an entry correctly predicted if the correct word is among the top 5 most probable predictions.

\subsection{Results}
\label{sec:results}
We split results according to the task into four subsections: NER, POS, DP, and WA. In the last subsection we attempt to draw conclusions related to the vocabulary size.

\subsubsection{NER results}
The results of the NER task are shown in Table~\ref{tab:results-monolingual-ner}. 
We report macro average $F_1$ scores of the three named entity classes. In summary, our Est-RoBERTa performs the best for Estonian, and our LitLat BERT performs the best for Lithuanian and Latvian. We present language specific observations below.

\begin{table}[!!htb]
\begin{center}
\caption{The results of NER evaluation task for various BERT models. The scores are macro average $F_1$ scores of the three named entity classes.} 
\label{tab:results-monolingual-ner}
\begin{tabular}{lccc}
\hline
Model & Estonian & Latvian & Lithuanian \\
\hline
mBERT & 0.901 & 0.849 & 0.809  \\
XLM-R & 0.907 & 0.867 & 0.793 \\
Est-BERT & 0.870 & - & - \\
FinEst BERT & 0.925 & - & - \\
LVBERT & - & 0.780 & - \\
LitBERTa & - & - & 0.630 \\  \hline
LitLat BERT & - & \textbf{0.875} & \textbf{0.847} \\
Est-RoBERTa & \textbf{0.928} & - & - \\
\hline
\end{tabular}
\end{center}
\end{table}

While Est-RoBERTa performs best on the Estonian NER, Est-BERT has a larger vocabulary size than Est-RoBERTa, but it was trained on a much smaller corpus. In comparison with multilingual FinEst BERT, Est-BERT was trained on more than twice as large Estonian corpus, but the size of its whole training set was less than the third of the FinEst BERT. FinEst BERT performs significantly better than Est-BERT and almost as good as Est-RoBERTa. 

We observe similar behaviour on Latvian. LVBERT and LitLat BERT were both trained on 0.5 billion token large Latvian corpus, but LitLat BERT significantly outperforms LVBERT. LitLat BERT was pre-trained also on Latvian and English corpora, thus on a larger overall training set.
XLM-R was trained on an even larger total dataset and used larger Latvian corpus than both LVBERT and LitLat BERT. However, XLM-R performs worse than LitLat BERT. The reason might be that XLM-R was trained on 100 languages, therefore its vocabulary for Latvian has to be significantly smaller compared to trilingual LitLat BERT, which is an overall winner on Latvian.
The same relation  between XLM-R and FinEst BERT is observed on Estonian: FinEst BERT outperforms XLM-R, despite the latter being trained on a larger corpus with more Estonian data.

Similar relation between massively multilingual, trilingual, and inferior monolingual models can be observed for Lithuanian. The mBERT model performs slightly better than XLM-R, but both lag behind LitLat BERT. Monolingual LitBERTa is significant worse than the other three models, but we do not have enough information about its training to explain its behaviour.

\subsubsection{POS-tagging results}
The results of the POS-tagging task (Table~\ref{tab:results-monolingual-pos}) show  similar behaviour as in the NER task. All the models score well on this task, especially XLM-R closes the gap to other models. On Lithuanian, XLM-R performs the best, narrowly beating LitLat BERT. LitLat BERT performs the best on Latvian, and Est-RoBERTa performs the best on Estonian. 

\begin{table}[!htb]
\begin{center}
\caption{The results of the POS-tagging evaluation task for various BERT models. The scores are micro average $F_1$ scores.}
\label{tab:results-monolingual-pos}
\begin{tabular}{lccc}
\hline
Model & Estonian & Latvian & Lithuanian \\
\hline
mBERT & 0.966 & 0.946 & 0.934  \\
XLM-R & 0.970 & 0.960 & \textbf{0.964} \\
Est-BERT & 0.961 & - & - \\
FinEst BERT & 0.973 & - & - \\
LVBERT & - & 0.945 & - \\
LitBERTa & - & - & 0.916 \\  \hline
LitLat BERT & - & \textbf{0.966} & 0.961 \\
Est-RoBERTa & \textbf{0.977} & - & - \\
\hline
\end{tabular}
\end{center}
\end{table}

\subsubsection{DP results}
The results of the DP task are shown in Table~\ref{tab:results-monolingual-dp}. Multilingual models, especially massive multilingual models perform significantly worse on this task. Unlike on the NER and POS-tagging task, all monolingual models perform very well here. The gap between monolingual and trilingual models is not large, though, especially between LitBERTa and LitLat BERT. The difference between Est-BERT and Est-RoBERTa suggests that training on more data offers significant improvement in performance, but only if the model is monolingual. XLM-R performs poorly, despite being trained on a larger Latvian and Lithuanian corpora than any other model.

\begin{table}[!htb]
\begin{center}
\caption{The results of DP evaluation task for various BERT models. The results are reported as unlabelled attachment score (UAS) and labelled attachment score (LAS).}
\label{tab:results-monolingual-dp}
\begin{tabular}{lcccccc}
\hline
 & \multicolumn{2}{c}{Estonian} & \multicolumn{2}{c}{Latvian} & \multicolumn{2}{c}{Lithuanian} \\
Model & UAS & LAS & UAS & LAS & UAS & LAS \\
\hline
mBERT & 66.9 & 56.3 & 65.4 & 54.6 & 56.1 & 44.1  \\
XLM-R & 76.2 & 68.9 & 76.5 & 69.3 & 65.0 & 56.6 \\
Est-BERT & 82.1 & 77.4 & - & - & - & - \\
FinEst BERT & 80.9 & 75.7 & - & - & - & - \\
LVBERT & - & - & \textbf{80.8} & \textbf{75.7} & - & - \\
LitBERTa & - & - & - & - & 68.6 & \textbf{62.6} \\  \hline
LitLat BERT & - & - & 80.4 & 74.3 & \textbf{68.7} & 61.7 \\
Est-RoBERTa & \textbf{83.2} & \textbf{78.6} & - & - & - & - \\
\hline
\end{tabular}
\end{center}
\end{table}
%\vspace{-10mm}
\subsubsection{WA results}
We were solving the word analogy task by predicting masked tokens. Because massively multilingual models split words into several subword tokens more frequently than monolingual models, we do not expect them to perform well here, as they need to be able to correctly predict a few consecutive masked tokens. The results in Table~\ref{tab:results-monolingual-analogy} confirm our expectation regarding the mBERT model. However, XLM-R performs well and it achieved the second best score in each language. Est-RoBERTa and LitLat BERT significantly outperform all the other models on this task. Surprisingly, LitBERTa performs poorly, although its vocabulary is large. 

\begin{table}[h!!tb]
\begin{center}
\caption{The results of word analogy task, solved as a masked token prediction task for various BERT models. The results are reported as macro average P@5 score.}
\label{tab:results-monolingual-analogy}
\begin{tabular}{lccc}
\hline
Model & Estonian & Latvian & Lithuanian \\
\hline
mBERT & 0.093 & 0.026 & 0.036  \\
XLM-R & 0.251 & 0.118 & 0.107 \\
Est-BERT & 0.165 & - & - \\
FinEst BERT & 0.224 & - & - \\
LVBERT & - & 0.118 & - \\
LitBERTa & - & - & 0.044 \\  \hline
LitLat BERT & - & \textbf{0.170} & \textbf{0.214} \\
Est-RoBERTa & \textbf{0.393} & - & - \\
\hline
\end{tabular}
\end{center}
\end{table}

\subsubsection{Analysis of vocabulary size}

\begin{figure}[h!!tb]
\includegraphics[width=\textwidth]{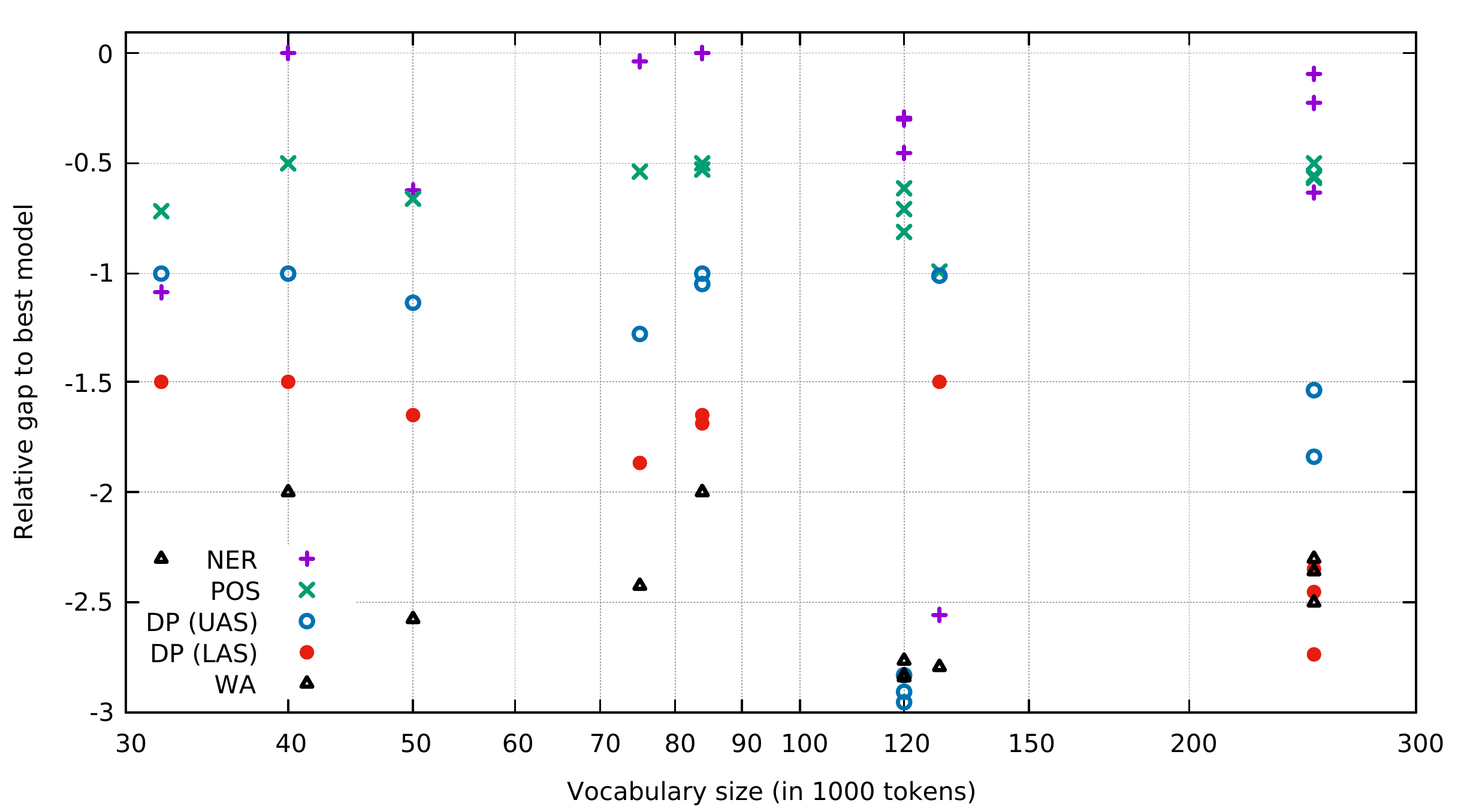}
\caption{The results expressed as a relative gap to the best model for each language per task: $1-\frac{x}{x_{best}}$. The gaps are increased ten-fold for all tasks, except WA, for better readability. The results are offset by task along the y-axis: NER by 0, POS by -0.5, DP (UAS) by -1.0, DP (LAS) by -1.5, WA by -2.0. } \label{fig1}
\end{figure}

In Figure~\ref{fig1}, we plot the relative performance gap of each model to the best model for that language against the model's vocabulary size. We observe, that for monolingual models, increasing the vocabulary size at first improves the performance. Further increasing the vocabulary size decreases the model's performance. The exception is the DP task, where larger vocabulary appears to be particularly important. In multilingual models, the larger vocabulary also improves the performance on all tasks. However, the results are strongly correlated with the training dataset sizes (not shown).
%\subsection{Zero-shot cross-lingual evaluation(?)}
% should we even include this?

 To better understand the performance of large pretrained models, we require a joint analysis of both, the training data size and the vocabulary size. As such a study would require building several large  models and therefore considerable computational resources, we leave it for further work.

\section{Conclusions}
\label{conclusions}
We trained and publicly released two new large pretrained masked language models, a monolingual Estonian Est-RoBERTa, and a trilingual LitLat BERT, trained on Latvian, Lithuanian, and English. Est-RoBERTa improves performance over existing BERT-like models for Estonian on all four evaluation tasks: NER, POS-tagging, DP, and WA. LitLat BERT outperforms related models for Latvian and Lithuanian on three tasks: NER, POS-tagging, and WA. On the DP task it offers comparable performance to the monolingual models, especially on Lithuanian, while outperforming multilingual models.

We have observed that training on more data does not necessarily bring better performance, as XLM-R lags behind monolingual models trained on relatively small corpora. However, a monolingual model's performance can be improved by adding additional training data in other languages. FinEst BERT  outperforms Est-BERT, even though it was trained on a much smaller Estonian dataset. Similarly, LitLat BERT and LVBERT were trained on Latvian corpora of similar size, but LitLat BERT significantly outperforms LVBERT on most tasks. The exact benefit of training a model on additional data in other languages over a monolingual model depends on the amount of monolingual data available, and on the evaluation task. On the DP task, a monolingual model is preferred, while on the POS-tagging task, there is little penalty for massively multilingual models.

Despite large requirements for computational resources, it would be interesting to systematically test the main parameters of large pretrained language models such as the number of languages, size of the dictionary, and similarity of included languages.
%Ignoring the particularly badly performing mBERT model (\num{120000} tokens), a few trends
%We do not observe any correlation between the size of the model's dictionary and the performance of the tested models. The results suggest that a large dictionary might be beneficial in tasks like DP, but detrimental in tasks like word analogy. Further study is needed to properly evaluate the influence of the dictionary size. \hl{Is this true? XLM-R compared to LitLAt BERT on Latvian might show that dictionary size is relevant.}

% No correlation between dictionary/vocab size and performance
% No correlation between training size and performance when comparing mono and multilingual models
% More data always better for monolingual models, larger dictionary not necessarily
% Extra data from other languages sometimes beneficial, but only up to a certain point, then it's detrimental
% Thorough, methodical evaluation needed to properly determine the influence of dataset size and dictionary size
% Monolinguality vital for dependency parsing; little importance in POS-tagging; NER and WA are a mixed bag.

\section*{Acknowledgements}
%Removed due to anonymity.

This paper is supported by European Union's Horizon 2020 research and  innovation programme under grant agreement No 825153, project EMBEDDIA (Cross-Lingual Embeddings for Less-Represented Languages in European News Media).
The results of this publication reflect only the authors' view and the EU Commission is not responsible for any use that may be made of the information it contains.
The work was partially supported by the Slovenian Research Agency (ARRS) through the core research programme P6-0411. 
\bibliographystyle{splncs04}
\bibliography{main}

\end{document}